\pdfoutput=1

\documentclass[11pt]{article}

\usepackage{acl}
\usepackage{graphicx}
\usepackage{booktabs}
\usepackage{times}
\usepackage{latexsym}
\usepackage{amsmath}
\usepackage{todonotes}
\usepackage[T1]{fontenc}

\usepackage[utf8]{inputenc}

\usepackage{microtype}
\usepackage{CJKutf8}
\title{SMoA: Sparse Mixture of Adapters to Mitigate Multiple Dataset Biases}

\author{\textbf{
Yanchen Liu\thanks{$^\ast$Equal contribution.\:\:Work done during Yanchen Liu's internship at Baidu. $^\dagger$Correspondence to: Jing Liu <liujing46@baidu.com>} $^{\ 1,\, 2}$, 
Jing Yan$^{\ast\ 2}$, 
Yan Chen$^{\ast\ 2}$, 
Jing Liu$^{\dagger\ 2}$,
Hua Wu$^{2}$
} \\
    $^1$ Harvard University
    $^2$ Baidu Inc.\\
    yanchenliu@fas.harvard.edu\\
    yanjing09, chenyan22, liujing46, wu\_hua@baidu.com
}

\begin{document}
\maketitle
\begin{abstract}
Recent studies reveal that various biases exist in different NLP tasks, and over-reliance on biases results in models' poor generalization ability and low adversarial robustness. To mitigate datasets biases, previous works propose lots of debiasing techniques to tackle specific biases, which perform well on respective adversarial sets but fail to mitigate other biases. In this paper, we propose a new debiasing method Sparse Mixture-of-Adapters (\textsc{SMoA}), which can mitigate multiple dataset biases effectively and efficiently. Experiments on Natural Language Inference and Paraphrase Identification tasks demonstrate that \textsc{SMoA} outperforms full-finetuning, adapter tuning baselines, and prior strong debiasing methods. Further analysis indicates the interpretability of SMoA that sub-adapter can capture specific pattern from the training data and specialize to handle specific bias.


\end{abstract}

\begin{CJK*}{UTF8}{gbsn}

\section{Introduction}\label{intro}
Recent studies demonstrate that there are various biases in existing datasets, across different tasks, such as Natural Language Inference (NLI)~\cite{mnli-hard, snli-hard, hans, lms, hyponli}, Paraphrase Identification (ParaI)~\cite{paws, apt, LLS} and Machine Reading Comprehension (MRC)~\cite{mrc_benchmark}. Models 
trained on these biased datasets tend to exploit shallow patterns rather than understand tasks. Over-reliance on biases results in models' poor generalization ability and low adversarial robustness~\cite{geirhos2020shortcut}.

To mitigate datasets biases, previous work have proposed debiasing techniques on various NLP tasks. 
Existing debiasing approaches could be divided into two groups, data-based and model-based. Data-based approaches mitigate biases with adversarial data augmentation~\cite{min2020syntactic} or data filtering~\cite{le2020adversarial}. To balance datasets' distribution, these approaches will add or delete training data which are against or contain a specific bias. Model-based approaches weaken the impact of biases via re-weighting training examples given by bias model or deal with specific bias by ensembling a bias-only model with the main model~\cite{clark-etal-2019-dont, he-etal-2019-unlearn, karimi-mahabadi-etal-2020-end}. Besides, some model-based approaches learn less biased representation for input to handle with biases~\cite{clark-etal-2019-dont, he-etal-2019-unlearn, karimi-mahabadi-etal-2020-end}.

However, most of the debiasing approaches can tackle only one specific bias, but are not effective when mitigating multiple biases at the same time \cite{nli_benchmark}, i.e. improve performance on one specific adversarial set but are not effective on other adversarial sets, or even harm the generalization ability of the model (see in Sec.~\ref{experiments sec}). Besides, some works ensemble different biased models to improve the bias models' generalization ability~\cite{wu2020improving,nli_benchmark}. But model ensemble method consumes multiplied computing resources so that it is not efficient in real-world applications. It is meaningful to design a debiasing technique which is effective and efficient to handle with various biases of different dimensions. 

\begin{figure}
  \includegraphics[width=\linewidth]{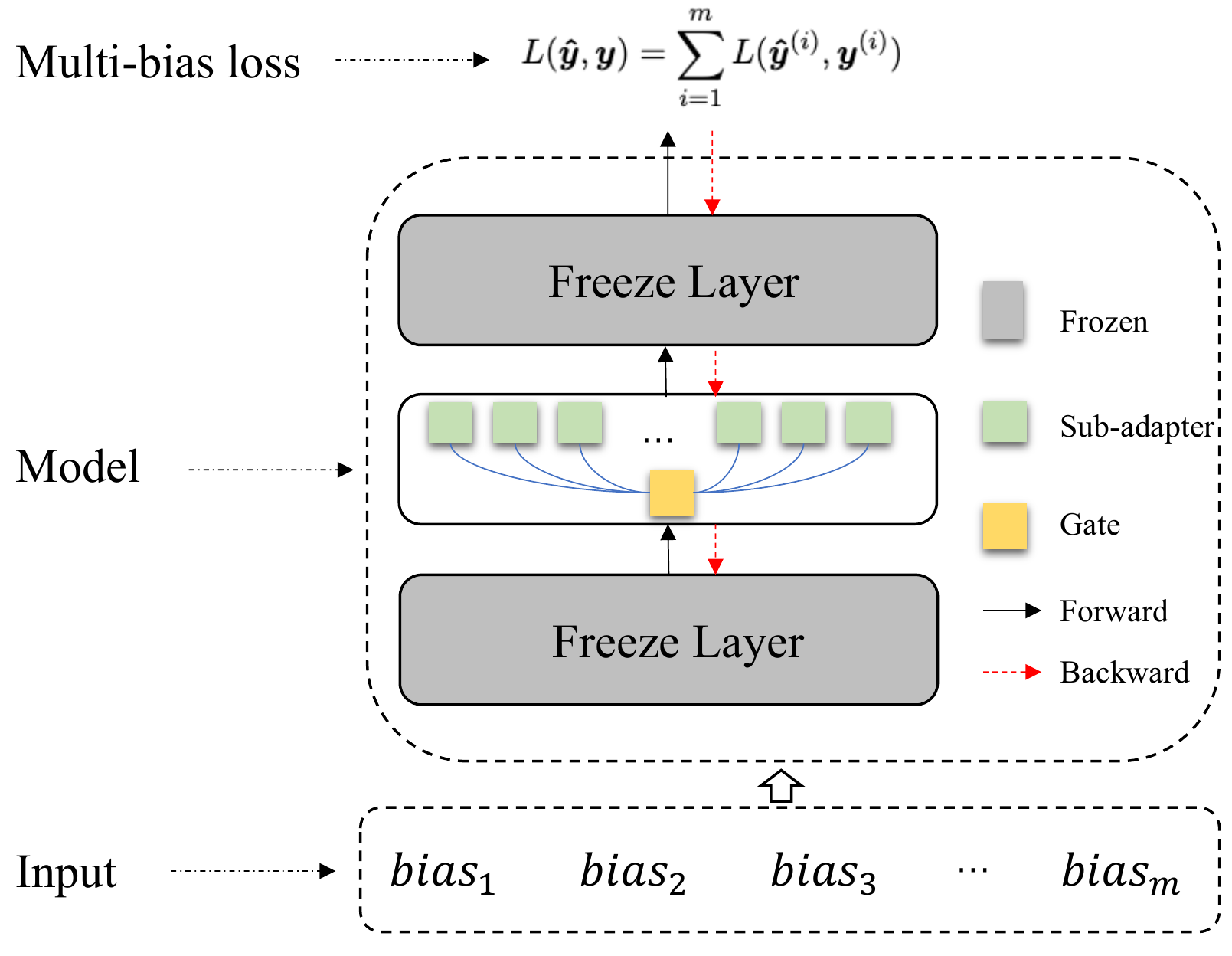}
  \caption{Illustration of overall debiasing framework.}
  \label{debiasing_framework}
\end{figure}

In this paper, we propose a debiasing framework to mitigate multiple dataset biases in effective and efficient way. Specifically, 1) by \textit{effective}, we design a framework to deal with multiple biases at the same time, and improve the performance on all adversarial sets same as or better than only tuning with its own adversarial training data; 2) by \textit{efficient}, the computational cost is lower than model ensemble method. Towards this end, we propose \textit{Sparse Mixture-of-Adapters (SMoA)} based on adapter modules proposed by~\cite{adapter}, which retains original parameters of the back-bone model fixed and supports multiple debiasing-parameters infused into the back-bone (illustrated in Fig~\ref{debiasing_framework}).
We consider dealing with multiple biases of different dimensions in one task at the same time a similar scenario as multi-task learning, and refer to as multi-bias mitigating. To mitigate multiple biases effectively, we insert multiple sub-adapters into backbone model to learn debiasing knowledge against different biases. 
The sub-adapters are treated as specialized experts to handle specific bias. SMoA outputs different representations about different types of debiasing knowledge without tuning the backbone model, which will retain the previously learned information upon learning new information and guarantee the model generalization ability. Besides, to be efficient in real-world applications, we employ \textit{sparse-gate} to select sub-adapters, which consumes the same computational cost as \textit{Adapter} method in the inference phase. Compared with tuning the entire backbone model, only $\sim$3.57\% of total parameters are tuned with SMoA (only the selected sub-adapters and the sparse gates are tuned with the mixture of adversarial sets for different biases).

Taking two of the most studied NLP tasks - NLI and ParaI - as examples, our experiments show that SMoA can improve the model robustness in multiple bias dimensions. And our method outperforms previous debiasing approaches with tuning only approximately 3.57\% of total parameters (see Sec.~\ref{experiments sec}). In summary, our contributions are as follows:
\begin{itemize}
  \item [1)] We propose a new framework \textsc{SMoA}, which can deal with multiple different biases simultaneously by learning from multiple adversarial datasets (see in Sec.~\ref{model sec}). 
  \item [2)] Experiments demonstrate the effectiveness and efficiency of SMoA. For NLI task, SMoA outperforms two-stages full finetuning by 1.22\% on average (by up to 1.08\% for hypothesis only bias \cite{hyponli}, 0.61\% for inter-sentences bias \cite{hans} and 3.04\% for lexical features only bias \cite{lms}), while only tuning about 3.57\% parameters. Similar results are shown on Paraphrase Identification task. The analysis of model prediction behaviors indicates that our trained sub-adapters can specialize to handle specific bias (see in Sec.~\ref{experiments sec}).
\end{itemize}


The remaining of this paper is organized as follows: in Sec.~\ref{model sec}, we propose SMoA debiasing method. In Sec.~\ref{experiments sec}, we experiment and analyse the effectiveness and efficiency of SMoA, and compare SMoA with other debiasing methods. In Sec.~\ref{related-work sec} and Sec.~\ref{conclusion sec}, we discuss related works and conclude our work.

\section{Sparse Mixture of Adapters}\label{model sec}

In this section, we firstly introduce 3 common types of biases in Natural Language Understanding (NLU) tasks. 
Then we propose \textit{Sparse Mixture-of-Adapters (SMoA)}, which is designed to mitigate multiple biases effectively and efficiently.

\subsection{Biases in NLU Tasks}
Intuitively, model trained on a basic dataset has basic competency to solve the basic NLP task, but may be deficient in some fine-grained sub-competencies, which leads to poor performance in out-of-domain or adversarial robustness dataset. The deficiency of sub-competency always shows up as a bias and could be seen as a "Gordian Knot" in a NLP task. We investigate the usual biases in NLU tasks and classify them as following three types:

\begin{itemize}
  \item [1)] \textbf{Lexical feature bias}: As showed by previous works \cite{lms, competency, LLS}, spurious correlations between word and label exist in many datasets. Models finetuned on these datasets tend to over-rely on examples' lexical features rather than understand tasks, and assign example with label highly correlated to a specific biased word in this example without understanding sentence meaning.
  
  \item [2)] \textbf{Partial-input only bias}: Previous work show that a large amount of examples in the dataset can be solved with only partial-input. \citet{snli-hard, hyponli} find that models can perform surprisingly well with only hypothesis accessible on SNLI~\cite{snli} and MNLI~\cite{mnli}, and refer this phenomenon as \textit{hypothesis-only bias}. Besides, some works show that model for Machine Reading Comprehension (MRC) exploits similar heuristic. \cite{mrc_benchmark} show that most of the questions in MRC datasets already answered correctly by the model do not necessarily require grammatical and complex reasoning, but only rely on content words or partial context.
  
  \item [3)] \textbf{Inter-sentences bias}: Model's over-reliance on inter-sentences overlap heuristic have been widely studied in sentence (text) pair tasks. \citet{hans} demonstrate that model trained on MNLI \cite{mnli} may adopt three fallible syntactic heuristics: lexical overlap heuristic, sub-sequence heuristic, the constituent heuristic. Similar heuristics have also been found by \cite{paws} in Paraphrase Identification task. They showed that model often predicts "Paraphrase" for a sentence pair based on high lexical overlap.
  
\end{itemize}



\subsection{Sparse Mixture-of-Adapters}\label{sec:SMoA_framework}
\paragraph{Model architecture.}
Recent works demonstrate that \textit{Parameter-Efficient Tuning} (PET) brings significant improvements in multi-task learning scenario~\cite{adapterfusion}. Dealing with different biases simultaneously can be seen as a similar scenario as multi-task learning, and we refer it as \textit{multi-bias mitigating}. To mitigate multi-bias at the same time, we propose \textit{Sparse Mixture-of-Adapters (SMoA)}, in which mixture-of-adapter layer (consist of $n$ sub-adapters) is inserted after each multi-head attention and feed-forward network layer, and it is controlled by sparse gating network~\cite{sparse-gate}.

\begin{figure}
  \includegraphics[width=\linewidth]{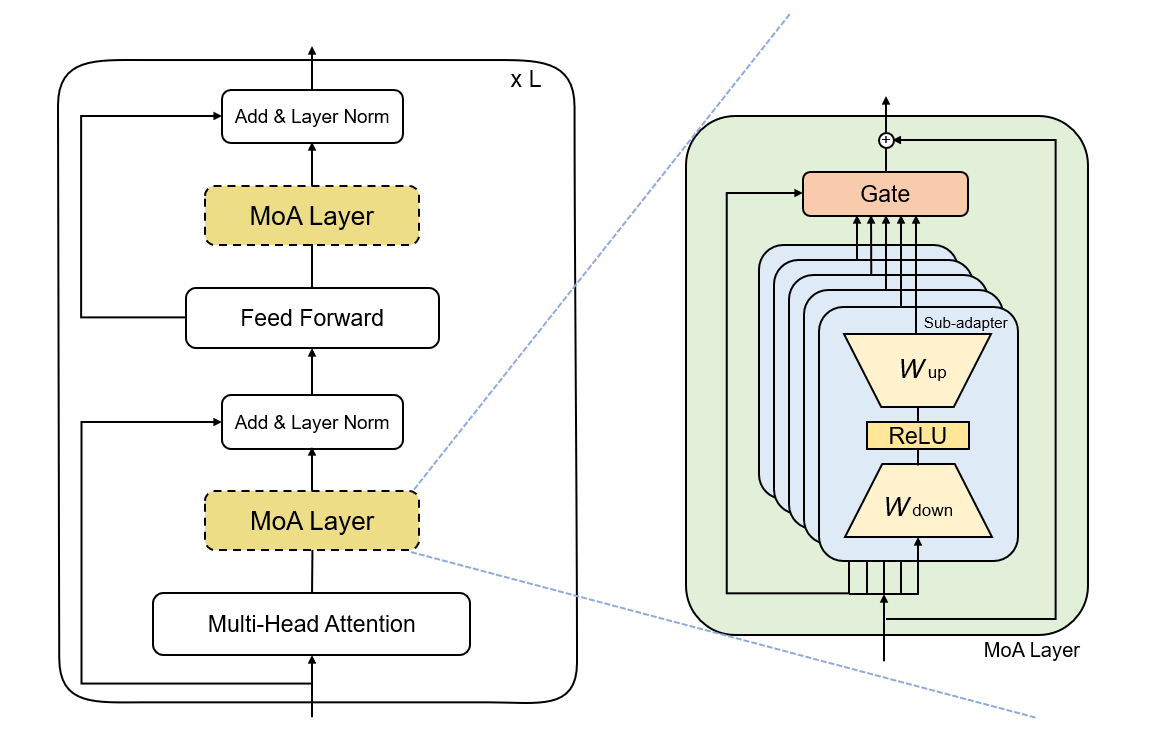}
  \caption{\textsc{SMoA} model architecture.}
  \label{SMoA}
\end{figure}

The model architecture is illustrated in Fig.~\ref{SMoA}. 
In a transformer-based backbone model\cite{attention}, a \textsc{SMoA} layer $\mathcal{A}$ is inserted after each multi-head attention and feed-forward network layer. 
\textsc{SMoA} layer $\mathcal{A}$ consists of $n$ sub-adapters ($\mathcal{A} = \left \{  a_1, a_2, ..., a_n \right \} $), and the sub-adapters are controlled by a sparse gating network. 
Compared to non-sparse gating~\cite{non-sparse}, after multiplying the input $x$ by a trainable weight matrix $W_g$, a sparse gating network keeps only the top $k$ values to guarantee the sparsity, and then will apply the softmax function $\sigma(\cdot)$~\cite{softmax}, which can denoted as follows:


\begin{equation}
    G(x) = \sigma(topK(x W_g ,k))
\end{equation}
where 
\begin{equation}
topK(v ,k)_i = 
\begin{cases}
 v_i & \text{ if } v_i \text{ is top } k \text{ element of } v,\\
 -\infty & otherwise.
\end{cases}
\end{equation}

As for a SMoA layer, for output $x$ from previous layer, the sparse gating network computes the weight for each sub-adapter and only keeps the top $k$ values, which will be multiplied with the outputs of sub-adapters $a_i(x)$, and then apply the softmax function. The \textsc{SMoA} layer's output can be denoted as 
\begin{equation}
O_{SMoA} = \sum_{i=1}^{n} G(x)_ia_i(x)
\end{equation}
where $n$ is the number of sub-adapters. Then the output will be passed into next layer with Add\&Norm \cite{resnet, LayerN}. 
\footnote{In practice, if there are $m$ adversarial datasets against $m$ biases respectively, we set the number of sub-adapters as $2m-1$ and $k$ as 2 (if $m$ is more than 2).}

\paragraph{Multi-bias loss.}
Similar as multi-task learning, the loss function of \textit{multi-bias mitigating} is defined as 
\begin{equation}
L(\boldsymbol {\hat{y}}, \boldsymbol {y}) = \sum_{i=1}^{m}L(\boldsymbol {\hat{y}}^{(i)}, \boldsymbol {y}^{(i)} ) 
\label{loss}
\end{equation}
where $m$ adversarial datasets against $m$ biases are learned. $\boldsymbol {y}^{(i)}$ and $\boldsymbol {\hat{y}}^{(i)}$ represent labels and predictions for $m$-th datasets respectively.

\paragraph{Debiasing Training.}
We take the on basic datasets finetuned model as backbone model $M$.
During training phase, we freeze the $M$'s parameters, and only tune the inserted \textsc{SMoA} layers with the mixture of adversarial datasets against different biases.
Intuitively, only training a small part while keeping the main model frozen could avoid substantial change of model's parameters and retain knowledge from previous basic datasets as much as possible. Besides, training different parts for different biases is able to deal with multiple biases at the same time. 
\section{Experiments}\label{experiments sec}
To examine SMoA's effectiveness and efficiency, we apply this framework to NLI and ParaI tasks, which are two of the most studied NLU tasks. We firstly consider NLI task, where several works about bias analysis and debiasing methods have been studied and proposed. In Sec.~\ref{PI}, we conduct experiment on ParaI task, which has been proven suffering from inter-sentences bias by previous works. In Sec.~\ref{efficiency}, we will discuss parameter-efficiency of SMoA. In Sec.~\ref{sec: SMoA_behavior_ana}, we conduct a behavior analysis to examine how sub-adapters works to handle different biases. In Sec.~\ref{sec: comparison}, we compare SMoA with other debiasing methods.

\subsection{Experimental Setup}\label{sec:experimental setup}
\paragraph{Baselines} There are commonly two ways tuning the model for mitigating biases with adversarial data: 
one way is to further finetune the on basic training set finetuned model with adversarial data for a specific bias (or sequentially finetuning with several adversarial datasets to deal with multiple biases), we refer it as \textsc{Two Stages Learning}; another way is to directly finetune the model with training dataset mixed by basic and adversarial data, we refer it as \textsc{One Stage Learning}. We conduct comparative experiments of the two ways.

\paragraph{Training Details}\label{appendix:train details}
We use \textit{RoBERTa-base} as backbone model. For each task we will consider 3 different biases. We choose the number of sub-adapters $n$ to be 5, and keep top 2 values for each \textsc {SMoA} layer. Model is trained with learning rate 0.0003, AdamW optimizer, and a linear learning scheduler for 5 epochs. We use accuracy as the evaluation metric.


\subsection{Natural Language Inference}\label{NLI}
In NLI task, a sentence pair of premise $p$ and hypothesis $h$ will be given, the goal is to decide whether or not $h$ is entailed by $p$ \cite{snli}. Previous works show that NLI models suffer from various biases existing in the datasets, such as hypothesis-only bias \cite{snli-hard, mnli-hard, hyponli}, inter-sentences bias \cite{hans}, lexical features bias \cite{lms, LLS}, etc. In this subsection, we will conduct experiments to examine if SMoA can mitigate different biases in NLI at the same time.

\subsubsection{Data} 
\paragraph{Training data.}
We choose MNLI \cite{mnli} as our basic dataset and 
use three adversarial datasets against different biases: \textit{HARD} (against \textbf{hypothesis-only bias}, filter out from MNLI training set), \textit{LLS} (against \textbf{lexical feature bias}, filter out from MNLI training set), and \textit{HANS}~\cite{hans} (against \textbf{inter-sentences bias}).

\paragraph{Evaluation data.} We evaluate models on NLI adversarial benchmark \cite{nli_benchmark} to obtain an overview of model's abilities, in which \textit{PI-CD} and \textit{PI-SP} for \textbf{hypothesis-only bias}, \textit{IS-SD} for \textbf{inter-sentences bias}, \textit{IS-CS} for \textbf{lexical features bias} and \textit{ST} for \textbf{inter-sentences} and \textbf{lexical features bias} especially. 

Dataset construction, statistics and details are provided in Appendix \ref{nli_train_data}.

\subsubsection{Experimental Results}\label{nli_training}

\begin{table*}[ht]
\centering
\resizebox{\textwidth}{!}{
\begin{tabular}{llllllllll} \toprule
                          & PI-CD    & PI-SP    & IS-SD    & IS-CS    & LI-LI    & LI-TS    & ST       & Avg.   & MNLI(id test)  \\ \hline
Roberta-base Baseline     & 75.87    & 79.51    & 71.53    & 71.8     & 89.33    & 85.04    & 72.43    & 77.93  & 87.44 \\ \hline
\multicolumn{10}{l}{\textbf{\textsc{Two Stages Learning}}}\\ 
full tuning           & 75.22	 & 83.29	& 86.33	   & 74.70	  & 86.11	 & 85.53	& 70.37	   & 80.22  & 85.06  \\
adapter tuning            & 76.17	 & 83.02	& 85.11	   & 74.54	  & \textbf{90.55}	 & 85.70	    & 71.05	   & 80.88  & 86.50  \\
\textsc {SMoA}              & 75.44	 & \textbf{84.37}	& \textbf{86.94}	   & \textbf{77.74}	  & 89.77	 & \textbf{85.70}	    & 70.11	   & \textbf{81.44} & 85.15\\  
\multicolumn{10}{l}{\textbf{\textsc{One Stage Learning}}}\\ 
full tuning          &\underline{76.39}	&81.40	&85.86	&73.17	&88.20	&\underline{85.68}	&71.79	&80.36 & \underline{87.47}\\
adapter tuning           &75.41	&81.13	&\underline{86.26}	&72.87	&\underline{90.40}	&85.24	&\textbf{72.75}	&80.58 & 87.42\\
\textsc {SMoA}             &\textbf{76.44}	&\underline{81.67}	&86.23	&\underline{74.54}	&89.17	&85.36	&\underline{72.45}	&\underline{80.84} & \textbf{87.87}\\
\multicolumn{10}{l}{\textbf{Ablation Experiment}}\\ (MNLI) + HARD & 75.04 & \textbf{85.98} & 74.67 &	\textbf{74.39} &	90.09 &	85.85 &	67.37 & 79.06 & 83.91 \\

(MNLI) + HANS & 72.71 & 78.98 & \textbf{87.26}	& 70.12	& 88.78	& 84.59	& 71.63 &	79.15 & 85.67 \\
(MNLI) + LLS &	74.98 &	77.09 &	75.49 &	72.10 &	87.62 &	84.60 &	69.94 & 77.40 & 86.56 \\
Model Ensemble & 74.89 & 81.40 & 77.56 & 73.02 & 89.44 & 85.85 & 70.67 & 78.98 & 86.87\\\bottomrule
\end{tabular}
}
\caption{Results of RoBERTa-base model on the NLI adversarial test benchmark proposed by \cite{nli_benchmark} and in-domain MNLI test set.}
\label{nli_benchmark_table_roberta}
\end{table*}
We conduct our experiment on \textit{RoBERTa-base} \cite{roberta}. As discussed in Sec. \ref{sec:experimental setup}, the model's learning of basic training set and adversarial sets can be implemented in two ways: \textsc{Two Stages Learning} and \textsc{One Stage Learning}. Tab. \ref{nli_benchmark_table_roberta} shows the experimental results of \textit{RoBERTa-base} on the NLI adversarial test benchmark and MNLI in-domain test set.

\paragraph{Two Stages Learning.}
With two stages learning, the pre-trained model \textit{RoBERTa-base} firstly learn the basic training dataset \textit{MNLI}, and the finetuned \textit{RoBERTa-base} is regarded as the backbone model.
Then three augmented datasets \textit{HARD}, \textit{LLS} and \textit{HANS} are mixed up and learned in three strategies: \textbf{full tuning} (tuning the entire RoBERTa-base model), \textbf{adapter tuning} (only tuning the inserted adapter layers), \textbf {SMoA} (only tuning the inserted \textsc {SMoA} layers
) with the mixed adversarial sets. 

As shown in \textsc{Two Stages Learning} of Tab.~\ref{nli_benchmark_table_roberta}, \textit {SMoA} outperforms baseline (only fine-tune \textit{RoBERTa-base} with MNLI training set), full tuning, and adapter tuning across almost all categories of NLI adversarial test benchmark. \textsc {SMoA} significantly outperforms baseline across all categories except for a slight decrease on \textit{PI-CD} and \textit{ST}, yielding an average 3.5\% gain; compared to full tuning and adapter tuning, \textsc {SMoA} shows higher improvements on \textit{PI-SP}, \textit{IS-SD} and \textit{IS-CS} sub-sections and an average 1.22\% and 0.56\% gains respectively. Especially, compared to full tuning, \textit {SMoA} shows improvement by 1.08\% on \textit{PI-SP} (hypothesis-only bias), 0.61\% on \textit{IS-SD} (inter-sentences bias) and 3.04\% on \textit{IS-CS} (lexical features bias). Compared to full tuning, SMoA yields an average 1.22\% gain (Column Avg.) with only tuning 3.57\% parameters (will be discussed in Sec. \ref{efficiency}), which demonstrates that \textit {SMoA} framework is able to tackle multiple biases at the same time on NLI task. 
\paragraph{One Stage Learning.}
We compare experimental results of \textsc{Two Stages Learning} with \textsc{One Stage Learning} (training with the original dataset and multiple adversarial datasets all at once). As shown in Tab. \ref{nli_benchmark_table_roberta}, for adapter tuning and \textsc {SMoA} on NLI adversarial benchmark, \textsc{Two Stages Learning} strategy outperforms \textsc{One Stage Learning}; for full tuning, \textsc{One Stage Learning} is slightly better.
Moreover, when using \textsc{One Stage Learning} strategy, \textsc {SMoA} brings considerable improvements (0.48\% and 0.25\%) compared to full tuning and adapter tuning. On the in-domain performance, as shown in Column MNLI, although \textsc{Two Stages Learning} brings an larger improvement on NLI adversarial test benchmark, SMoA with \textsc{One Stage Learning} demonstrates a better performance on in-domain MNLI test set.

\paragraph{Ablation experiment.}
To analysis the effect of each adversarial training set separately, we use \textit{HANS}, \textit{HARD} and \textit{LLS} datasets separately to full tune the entire MNLI trained model. Ablation Experiment of Tab. \ref{nli_benchmark_table_roberta} demonstrates that the augmented datasets bring significant improvements on their corresponding bias: training with \textit{HARD} brings 6.47\% improvement on \textit{PI-SP}
subsection, with \textit{HANS} 15.73\% on \textit{IS-SD} and with \textit{LLS} 0.3\% on \textit{IS-CS}. Surprisingly, training with \textbf{HARD} shows greater improvement on \textit{IS-CS} than \textit{LLS}, the specific augmented dataset aiming to lexical feature only bias. However, an improvement on the specific bias brings decreases on other biases. As shown in Column Avg. of Tab. \ref{nli_benchmark_table_roberta}, "single-tuning" performs bad on the average performance, and SMoA performs better generally.

Moreover, we ensemble above three full-tuned models and use most-voting to do prediction. \textit{Model Ensembling} shows a better ability on in-domain test set compared to "single-tuning", but the performance of ensemble model on adversarial test sets is not good.

\indent In general, SMoA can tackle multiple biases at the same time, which shows a significant performance on adversarial robustness evaluation sets with tuning much less parameters.

\subsection{Paraphrase Identification}\label{PI}
Furthermore, we consider Paraphrase Identification (ParaI) task, in which a sentence pair of $s1$ and $s2$ is given, the goal is to decide whether or not $s1$ and $s2$ have the same meaning. Previous works \cite{paws, apt} show that existing ParaI datasets
lack sentence pairs that have high lexical and syntactical overlap without being paraphrases as well as pairs lexically and syntactically disparate but being paraphrases. Therefore, models trained on these datasets tend to predict sentence pairs with high overlap as "Paraphrase", low overlap as "Non-paraphrase", which usually results in failure in real-world data. 
Besides, \cite{LLS} reports models of ParaI task suffer from \textit{lexical feature bias}, i.e. models' over-reliance on spurious relations between label and word.

\subsubsection{Data}
We consider inter-sentences and lexical feature biases for ParaI task.

\paragraph{Training Data.}For basic training, we choose \textit{Quora Question Pairs (QQP)} Dataset \footnote{https://data.quora.com/First-Quora-Dataset-Release-Question-Pairs}. For adversarial training, we use \textit{PAWS} (against \textbf{high-overlap bias}), \textit{APT} (against \textbf{low-overlap bias}) proposed by \cite{paws, apt} and \textit{LLS} (against \textbf{lexical feature bias}) to reduce models’ over-reliance on inter-sentences and lexical feature biases. 

\paragraph{Evaluation Data.}
We use \textit{PAWS\textsubscript{QQP}}, \textit{PAWS\textsubscript{wiki}} and \textit{LLS} as evaluation sets. As there is not available evaluation dataset for models' compositionality sensitivity in ParaI task, we filter out unbiased examples from \textit{QQP} training set as \textit{LLS} training set, and unbiased examples from \textit{QQP} test set as \textit{LLS} evaluation set \footnote{The definition of unbiased examples is from \cite{LLS}.} 

Data statistics of adversarial ParaI datasets are provided in Appendix \ref{parai_data}.


\subsubsection{Experimental Results}
We use on \textit{QQP} finetuned \textit{RoBERTa-base} \cite{roberta} model as the backbone model for our \textsc {SMoA} method and train with augmented dataset consisting of above three adversarial datasets (\textsc{Two Stages Learning}). In Sec. \ref{nli_training}, we compare our method with full tuning and adapter tuning.

\begin{table*}[]
\centering
\resizebox{\textwidth}{!}{
\begin{tabular}{lllllll} \toprule
\multicolumn{1}{c}{}                          & \multicolumn{1}{c}{PAWS\textsubscript{QQP}} & \multicolumn{1}{c}{PAWS\textsubscript{WIKI}}                    & \multicolumn{1}{c}{APT}                          & \multicolumn{1}{c}{LLS}                         &\multicolumn{1}{c}{Avg.}   & \multicolumn{1}{c}{QQP(id test)}                  \\
 \hline
RoBERTa-base Baseline          & 39.14& 47.11                           & 69.94                                & \textbf{91.49}                                    &61.92 & \textbf{91.47}                                   \\ \hline
\multicolumn{1}{l}{\textbf{\textsc{Two Stages Learning}}} &&&&&\\
full tuning                                         & 89.66 &              81.36                   &                 77.16                   &                    88.41                   &84.14 & \underline{88.19}                                  \\
adapter tuning                                        & \underline{91.73}&                  \underline{81.70}                                   & \underline{77.16}                   &                          \underline{88.98}           &\underline{84.89} & 87.28                                   \\
\textsc {SMoA}                                                       & \textbf{92.17}& \textbf{81.71}                                   & \textbf{78.27}                   &                       88.05                                 &\textbf{85.05}& 87.72                    \\
\bottomrule
\end{tabular}}

\caption{For ParaI task, we train on QQP finetuned model with augmented  datasets composed of PAWS\textsubscript{QQP}, APT and LLS train sets using three different strategies: full tuning and adapter tuning and \textsc {SMoA}. The table shows the accuracy on in-domain test set QQP, and PAWS\textsubscript{QQP}, APT, LLS, PAWS\textsubscript{WIKI} evaluation sets.}
\label{qqp_table}
\end{table*}

Tab. \ref{qqp_table} shows that compared to baseline, training with our constructed adversarial training data leads to significant improvements on \textit{PAWS\textsubscript{QQP}}, \textit{PAWS\textsubscript{WIKI}} and \textit{APT} aiming to \textit{high-overlap bias} and \textit{low-overlap bias} respectively. It is worth noting that the accuracy improvement on \textit{PAWS\textsubscript{QQP}} is up to 53.03\%. Besides, \textsc {SMoA} outperforms full tuning and adapter tuning on \textit{PAWS\textsubscript{QQP}} and \textit{ADV} by 0.44\% and 1.11\%. However, all three strategies do not perform well on in-domain \textit{QQP} test set and \textit{LLS} evaluation set. We guess the reason is that original QQP dataset does not suffer from \textit{lexical features bias} so much that \textit{LLS} evaluation set has a similar distribution with \textit{QQP} test set.

\subsection{Parameter Efficiency Analysis}\label{efficiency}
The experimental results have shown that SMoA can tackle multiple biases at the same time effectively. Here we discuss the parameter efficiency of SMoA. As shown in Tab. \ref{tab:efficiency}, only 3.70\% parameters are tuned with \textsc {SMoA} compared to full tuning. If comparing with debiasing method using model ensembling, the gains of \textit{parameter-efficiency} will be even more significant. 

\begin{table}[h!]
\centering
\resizebox{\linewidth}{!}{
\begin{tabular}{llll}\toprule
                        & SMoA       & BERT-base & Ratio  \\\hline
\#total parameters     & 129258939  &  124647939 & 1.0370 \\
\#trainable parameters & 4611000    &  124647939 & 0.0370 \\
trainable ratio        & 3.57\%     &  100\%     & 0.0357 \\
forward time (s)       & 0.2356     &  0.0863    & 2.7300 \\\bottomrule
\end{tabular}}
\caption{Comparison of the number of parameters and forward time of vanilla RoBERTa-base model and \textsc {SMoA} (with 5 sub-adapters). The inference time is tested using a single A100-SXM4-40GB.}
\label{tab:efficiency}
\end{table}

\subsection{Sub-adapters' Behavior Analysis}\label{sec: SMoA_behavior_ana}
\begin{figure}
  \includegraphics[width=\linewidth]{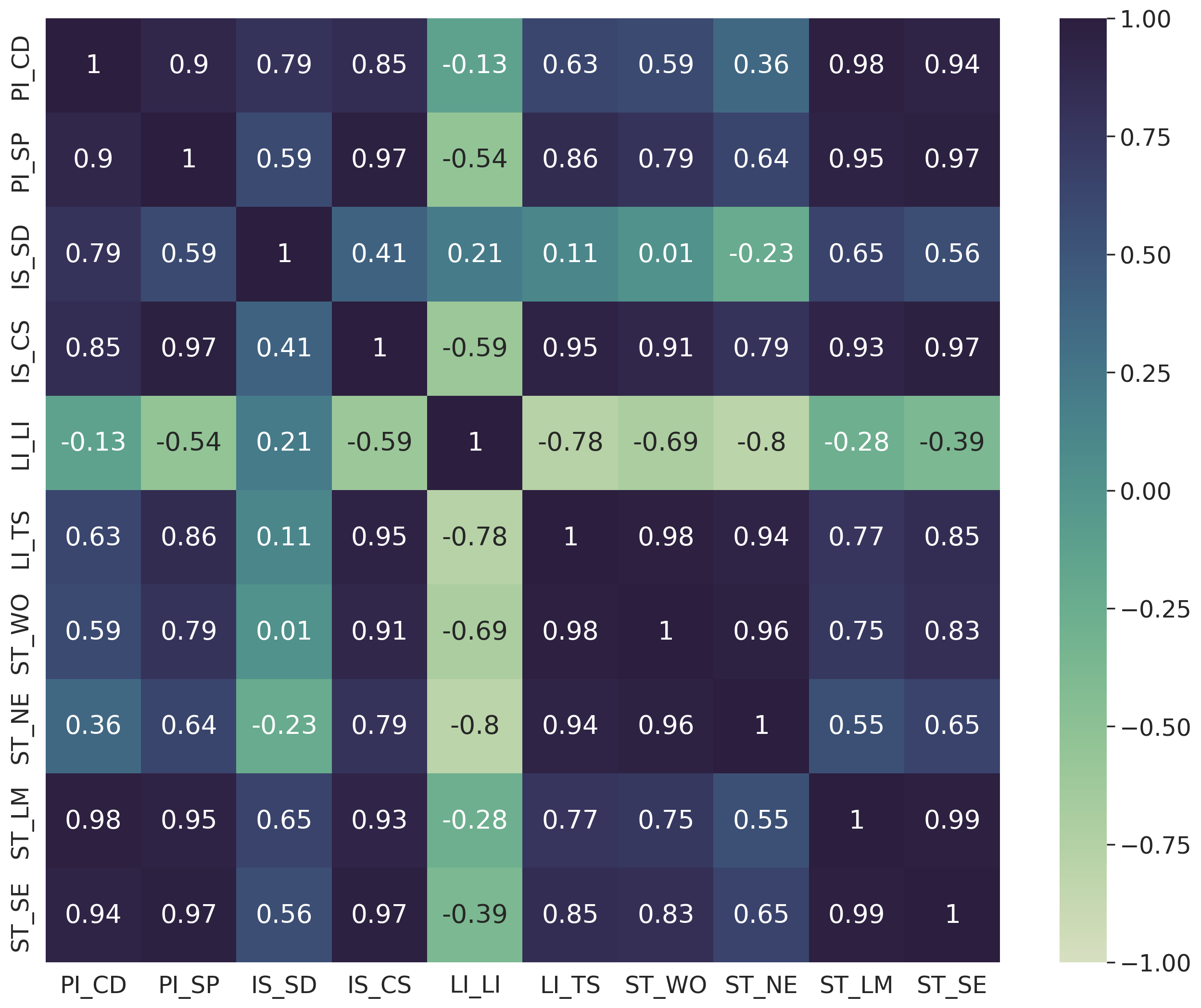}
  \caption{The Pearson’s correlation coefficients of top 2 sub-adapters distribution of the last \textsc {SMoA} layer between different
  adversarial subsets of the NLI adversarial benchmark \citep{nli_benchmark}. }
  \label{corr}
\end{figure}
We consider that each trained sub-adapter could capture specific patterns from the training data and learn the capability to handle specific bias. To verify this hypothesis, we analysis the distribution of the selected top 2 sub-adapters for each adversarial subset of the NLI adversarial benchmark \citep{nli_benchmark}. We calculate the distribution for each \textsc {SMoA} layer of our trained model in Sec. \ref{NLI} and find that the distribution for each adversarial subset are almost identical in the low layers. However, the distribution for each adversarial subset differs from each other in the high layers and demonstrates some correlations.

We plot the Pearson’s correlation coefficients of the distributions for the subsets in the last layer in Figure \ref{corr}. The distributions of selected sub-adapters are more correlated for the subsets with higher similarities. For example, the correlation coefficient is up to 0.9 for subset \textit{PI\_CD} and \textit{PI\_SP}, both are constructed with Partial-input (PI) heuristics. On the contrary, the correlation score of selected sub-adapters is low to -0.23 between \textit{IS\_SD} and \textit{ST\_NE}, which have surrogate correlation score of only 32.9/100 as reported by \citet{nli_benchmark}.

Test examples with the similar biases tend to be handled by the same set of sub-adapters, which implies SMoA does not only add more parameters, but could specialize to handle specific bias.

\subsection{Comparison with other Debiasing Methods}\label{sec: comparison}

\begin{table*}[ht]
\centering
\resizebox{\textwidth}{!}{
\begin{tabular}{llllllllll} \toprule
                          & PI-CD    & PI-SP    & IS-SD    & IS-CS    & LI-LI    & LI-TS    & ST       & Avg.   & MNLI  \\ \hline
BERT-base Baseline                  & $70.3_{\pm0.5}$ & $73.7_{\pm1.4}$ & $53.5_{\pm2.3}$ & $64.8_{\pm1.4}$ & $85.5_{\pm0.9}$ & $81.6_{\pm1.4}$ & $69.2_{\pm0.8}$ & $71.2_{\pm0.8}$ & 83.5\\ \hline
\multicolumn{10}{l}{\textbf{\textsc{Two Stages Learning}}}                             \\ 
full tuning           & 73.08    & 75.47    & 85.84    & 66.46    & 86.21    & 84.04    & 67.01    & 76.87  & 82.52 \\
adapter tuning            & 73.35    & 74.12    & 83.75    & 67.07    & 87.31    & 83.38    & 67.15    & 76.59  & 83.07 \\
\textsc {SMoA}              & 73.44    & 79.78    & 88.41    & 69.82    & 85.03    & 83.39    & 66.91    & 78.11  & 83.41 \\ \hline
\multicolumn{10}{l}{\textbf{Data-augmentation heuristics proposed by \cite{nli_benchmark}}} \\ 
Text Swap                & 71.7     & 72.8     & 63.5     & 67.4     & 86.3     & 86.8     & 66.5     & 73.6  & 83.7  \\
Sub (synonym)            & 69.8     & 72.0     & 62.4     & 65.8     & 85.2     & 82.8     & 64.3     & 71.8  & 83.5  \\
Sub (MLM)                & 71.0     & 72.8     & 64.4     & 65.9     & 85.6     & 83.3     & 64.9     & 72.6  & 83.6  \\
Paraphrase               & 72.1     & 74.6     & 66.5     & 66.4     & 85.7     & 83.1     & 64.8     & 73.3  & 83.7  \\ 

\multicolumn{10}{l}{\textbf{Prior debiasing strategies}}                                                                    \\
\cite{wu-etal-2022-generating} & $71.7_{\pm0.9}$ & $77.8_{\pm1.2}$ & $66.9_{\pm3.7}$ & $71.1_{\pm0.7}$ & $89.1_{\pm1.0}$ & $82.3_{\pm0.9}$ & $69.3_{\pm0.8}$ & $75.4_{\pm0.8}$ & 82.70\\\bottomrule
\end{tabular}}
\caption{Results on the adversarial benchmark proposed by \cite{nli_benchmark}. We compared our method with full tuning and adapter tuning, data augmentation as well as other prior debiasing strategies. Due to experiments' compatibility, we directly report the results from previous works.}
\label{nli_benchmark_table}
\end{table*}

To better compare the effectiveness of SMoA with previous proposed methods, we conduct experiments on \textit{BERT-base} for NLI task. As shown in table \ref{nli_benchmark_table}, when using a on MNLI finetuned \textit{BERT-base} model as backbone model and further training it on the adversarial datasets (\textsc{Two Stages Learning}),  \textsc {SMoA} significantly outperforms baseline across all categories except for a slight decrease on \textit{ST}, yielding an average 6.9\% gain. Compared to full tuning and adapter tuning, \textsc {SMoA} leads to better performance on \textit{PI-CD}, \textit{PI-SP}, \textit{IS-SD} and \textit{IS-CS} and an average 1.24\% and 1.5\% gains respectively. Especially, \textsc {SMoA} brings 4.31\% improvement on PI-CD (hypothesis-only bias), 2.57\% on IS-SD (inter-sentences bias) and 2.75\% on IS-CS (lexical features bias). Surprisingly, \textsc {SMoA}'s in-domain performance for \textit{BERT-base} model drop negligibly, that \textsc {SMoA} mitigates multiple biases at the same time while keeping model's in-domain capability.

In Tab. \ref{nli_benchmark_table}, we compare our method with various data-augmentation methods proposed by \cite{nli_benchmark}. As we can see, \textsc {SMoA} outperforms data-augmentation methods by 5.28\% in average. These data-augmentation methods are simple, and designed only for a single bias. \cite{wu-etal-2022-generating} proposed a data-generation-debiasing method to mitigate spurious correlations. Compared to this method, \textsc {SMoA} shows a better performance of 2.71\% on average, especially 1.74\% and 1.98\% improvements for \textit{hypothesis-only bias}. 
 
\section{Related Work}\label{related-work sec}

\paragraph{Bias Analysis and Evaluation.}
Spurious correlations in existing datasets have been studied across various NLP tasks. For example, previous work demonstrated that NLI system suffers from hypothesis-only bias \cite{snli-hard, hyponli}, inter-sentences bias \cite{hans} and lexical features bias \cite{lms}. For Paraphrase Identification, there are also biases have been widely studied \cite{paws, apt}. 

In addition, some work attempted to analyze the spurious correlations from a theoretical perspective \cite{competency, LLS}.  

\paragraph{Debiasing Methods.}
Previous debiasing methods tackled biases from either model level or dataset level. 

On model level,
\citet{DBLP:journals/corr/abs-2004-09034} introduce a new training objective to utilize counterfactual examples for debiasing, \citet{belinkov-etal-2019-dont, belinkov-etal-2019-adversarial, zhou-bansal-2020-towards} propose to learn less biased representation for input, \citet{clark-etal-2019-dont, he-etal-2019-unlearn, karimi-mahabadi-etal-2020-end} deal with specific bias by ensembling a bias-only
model with the main model or weaken the impact of biases via re-weight training examples given by a bias model.

On dataset level, \citet{DBLP:journals/corr/abs-1907-10641, DBLP:journals/corr/abs-2002-04108} propose to filter the existing dataset to obtain a debiased one. Besides, a series of data augmentation strategies proposed by \cite{DBLP:journals/corr/abs-2107-07150, DBLP:journals/corr/abs-2101-00288, wang-etal-2021-cline, DBLP:journals/corr/abs-2105-08206,DBLP:journals/corr/abs-2012-13985, nli_benchmark} also have been demonstrated to improve robustness. \citet{wu-etal-2022-generating} combine data augmentation with dataset filtering to mitigate the spurious correlations in existing datasets.

\paragraph{Parameter Efficient Learning.} Previous work demonstrated that Parameter Efficient Learning \cite{DBLP:journals/corr/abs-2110-04366, https://doi.org/10.48550/arxiv.2203.06904}, including prompt tuning \cite{prompttuning}, prefix tuning \cite{prefix} and adapter tuning \cite{adapter}, could achieve comparable and even better performance and robustness than full finetuning with only much fewer parameters \cite{https://doi.org/10.48550/arxiv.2110.07602, DBLP:journals/corr/abs-2110-04366, soup}.

\citet{adapterfusion} propose a framework based on adapter to incorporate knowledge from multiple tasks and achieve significant success in multi-task learning. ~\citet{wang2022adamix} proposes a stochastic routing strategy to ensemble adapters, which averages the adapter parameters for low-cost storage and inference. Similar strategy of ensembling multiple adapters are used in~\citet{Asai2022ATTEMPTPM},~\citet{Wang2020KAdapterIK} and ~\citet{pfeiffer2021adapterfusion}.
Previous work significantly improves models' performance on specific sets, but general good performance on all sets remains unexplored.
However, there is currently no work investigating whether Parameter efficient learning could be used to mitigate several different biases for one task simultaneously.

\section{Conclusion}\label{conclusion sec}
Most of the existing debiasing methods focus on tackling one specific bias, which leads to significant improvements on specific adversarial test sets, but poor performance on others. However, different biases exist at the same time in the real-world applications. To mitigate multiple biases effectively and efficiently, we propose Sparse Mixture-of-Adapters (\textsc{SMoA}). 
Experimental results on NLI and ParaI task demonstrate that \textsc{SMoA} outperforms previous debiasing methods across various biases. Furthermore, we examine the interpretability of SMoA that the sub-adapters in SMoA can specialize to handle specific bias. However, the added sparse gates and the selected sub-adapters cost more inference time, and SMoA architecture should be retrained if there are additional new bias types. In the future work, we will explore how to further improve the effect and efficiency of model debiasing method.

\end{CJK*}

\bibliography{anthology,moa}
\bibliographystyle{acl_natbib}

\begin{CJK*}{UTF8}{gbsn}
\appendix
\section{Dataset Details}\label{dataset_detail}
\subsection{Natural Language Inference}
\textbf{1. Adversarial training sets}\label{nli_train_data}

\textit{HARD} and \textit{LLS} are filter in MNLI training set, and \textit{HANS} is proposed in \cite{hans}. Data statistics are provided in Tab. \ref{nli_statistics}.

\begin{itemize}
  \item [1)] \textbf{HARD}: To against \textbf{hypothesis-only bias} in NLI, as \cite{snli-hard, mnli-hard}, we use a finetuned external model (\textit{RoBERTa-large}~\cite{roberta}) to filter the examples in original MNLI training set that are not correctly classified if only hypothesis is provided to construct a adversarial dataset \textit{HARD}.\footnote{In the original MNLI training set, there are many examples cannot be correctly classified with only hypothesis. We randomly select 40500 label balanced examples as train set and 4500 as dev set.}
  
  \item [2)] \textbf{LLS}: As \cite{LLS}, who define the word highly co-occurring with a specific label as \textit{biased word} of a dataset, we analysis the co-occurrence of words in MNLI training set and specific labels to obtain the \textit{biased words}, and filter out the examples not containing any biased words ("unbiased examples") as \textit{LLS}.\footnote{We choose the words with frequency $\geq$ 3 and most possible label ratio frequency $\geq$ 0.385 as biased word. We balance the unbiased examples set and split it as train and dev subset.} 
  We expect to mitigate model's over-reliance on the \textbf{lexical feature bias} by incorporating the debiased dataset \textit{LLS} into training.
  
  \item [3)] \textbf{HANS}: Except above two augmented datasets filtered from original training set, we consider an existing dataset \textit{HANS} \cite{hans}. \textit{HANS} provides a training set and an evaluation set. HANS evaluation set includes premise-hypothesis pairs with high lexical, sub-sequence and constituent overlap but not semantically entailable where model's overlay heuristics fail.
  We use \textit{HANS} training set, and aim to mitigating model's \textbf{inter-sentences bias}. 
  In their original training set, there are only two labels (entailment and non-entailment), because they think the distinction between contradiction and neutral was often unclear for their cases. In order to use this dataset for training a three-class classification model, we label the "non-entailment" subset with an external large finetuned model, filter out as "entailment" incorrectly classified examples, merge with "entailment" subset and balance the label distribution. 
  
\end{itemize}

\begin{table}[]
\centering
\begin{tabular}{l|lll}\toprule
\multicolumn{1}{c|}{} & \multicolumn{1}{c}{HARD} & \multicolumn{1}{c}{HANS} & \multicolumn{1}{c}{LLS} \\ \hline
train                & 40500                    & 3261                     & 37065                   \\
dev                  & 4500                     & 363                      & 4119                   
\\ \bottomrule \end{tabular}
\caption{Data statistics of adversarial datasets of NLI.}
\label{nli_statistics}
\end{table}

\textbf{2. Adversarial test sets}\label{nli_test_data}
We use NLI adversarial test benchmark~\cite{nli_benchmark} as test sets.
\begin{itemize}
  \item [1)] \textbf{Partial-input (PI) heuristics}: 
  
  \textit{PI-CD}: a subset of SNLI test set built by \cite{snli-hard} aiming to hypothesis-only bias, also known as SNLI-hard.
  
  \textit{PI-SP}: a subset of MultiNLI mismatched dev set built by \cite{hyponli} aiming to surface patterns heuristics in NLI.
  
  \item [2)] \textbf{Inter-sentences (IS) heuristics}: 
  
  \textit{IS-SD}: syntactic diagnostic dataset HANS \cite{hans}.
  
  \textit{IS-CS}: \cite{lms} compute the ’lexically misleading scores (LMS)’ for each instance in the SNLI test and MNLI dev sets using a softmax regression model to measure the importance of compositional information (not only lexical features) for solve this example. IS-CS is the subset whose LMS are larger that 0.7.
  
  \item [3)] \textbf{Logical-inference (LI) ability}: 
  
  \textit{LI-LI}: lexical inference test dataset by \cite{glockner-etal-2018-breaking, naik-etal-2018-stress}.
  
  \textit{LI-TS}: adversarial examples created by swapping the premise and hypothesis aiming to first-order logical inference ability \cite{minervini-riedel-2018-adversarially, DBLP:journals/corr/abs-1809-02719}.
  
  
  \item [4)] \textbf{Stress test (ST)}: An aggregation of “word overlap”, “negation” , “length mismatch” and “spelling errors” tests in \cite{naik-etal-2018-stress}. 
  
\end{itemize}

\subsection{Paraphrase Identification}\label{parai_data}
Tab.~\ref{qqp_statistics} presents the data statistics of adversarial datasets of ParaI task, PAWS\textsubscript{QQP}, APT and LLS. For PAWS\textsubscript{QQP} and APT, we split the original train sets as our train and dev subsets.

\begin{table}[ht!]
\centering
\begin{tabular}{l|lll} \toprule
\multicolumn{1}{c|}{} & \multicolumn{1}{c}{PAWS\textsubscript{QQP}} & \multicolumn{1}{c}{APT} & \multicolumn{1}{c}{LLS} \\ \hline
train                & 10788                         & 3370                    & 9000                    \\
dev                  & 1200                          & 376                     & 1000                    \\
test                 & 677                           & 1262                    & 1398                   
\\ \bottomrule \end{tabular}
\caption{Statistics of adversarial datasets of ParaI task.}
\label{qqp_statistics}
\end{table}
\end{CJK*}


\end{document}